\documentclass{article}

\PassOptionsToPackage{numbers, sort, compress}{natbib}

\newcommand{\isArXiv}[2]{#1}

\isArXiv{
\usepackage[preprint]{neurips_2019}
}{
\usepackage{neurips_2019}
}

\usepackage[utf8]{inputenc} %
\usepackage[T1]{fontenc}    %
\usepackage{hyperref}       %
\usepackage{url}            %
\usepackage{booktabs}       %
\usepackage{amsfonts}       %
\usepackage{nicefrac}       %
\usepackage{microtype}      %

\usepackage{graphicx}
\usepackage{caption}
\usepackage{subcaption}
\usepackage{amsmath}
\usepackage{booktabs}
\usepackage{comment}

\usepackage{epstopdf}
\usepackage{xspace}
\def\eg{\emph{e.g.}\@\xspace}
\def\ie{\emph{i.e.}\@\xspace}
\def\etal{\emph{et al.}\@\xspace}
\def\cf{\emph{c.f.}\@\xspace}
\def\etc{\emph{etc}}

\renewcommand{\paragraph}[1]{\noindent{\bf{#1}}}

\DeclareMathOperator*{\argmax}{arg\,max}
\newcommand{\loss}{\mathcal{L}}

\isArXiv{
\hypersetup{pdfauthor={Relja Arandjelovic},pdftitle={Object Discovery with a Copy-Pasting GAN}}
}{
}

\title{Object Discovery with a Copy-Pasting GAN}

\author{%
Relja Arandjelovi\'c$^\dagger$
\quad\quad\quad
Andrew Zisserman$^{\dagger,*}$ \\
$^\dagger$DeepMind
\quad
$^*$VGG, Department of Engineering Science, University of Oxford
}

\begin{document}

\maketitle

\newcommand{\figCpgan}{
\begin{figure}[t]
    \centering
    \begin{subfigure}[t]{0.32\linewidth}
    		\centering
        \includegraphics[height=7cm]{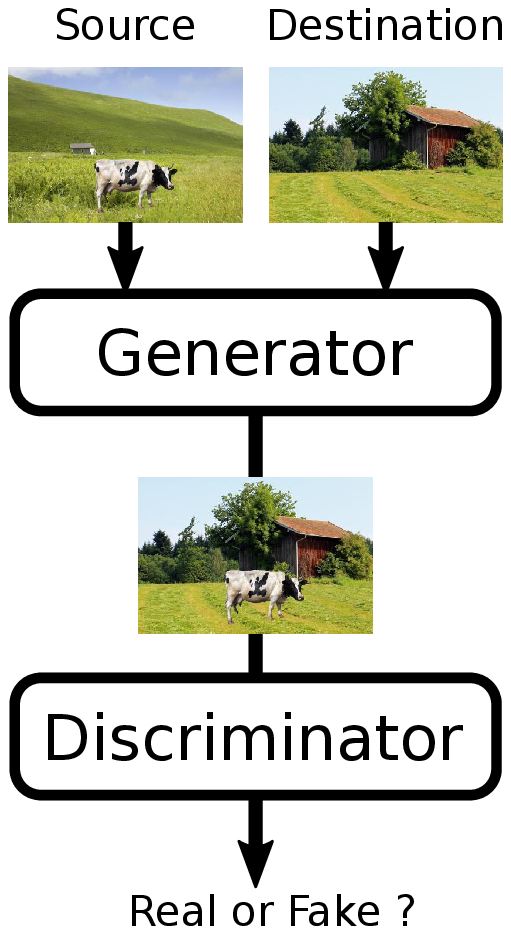}
        \caption{}
        \label{fig:cpgan:cpgan}
    \end{subfigure}
    \hfill
    \begin{subfigure}[t]{0.32\linewidth}
    		\centering
        \includegraphics[width=3.37cm]{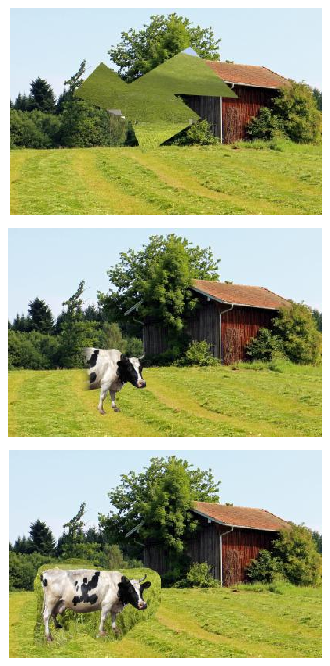}
        \caption{}
        \label{fig:cpgan:fake}
    \end{subfigure}
    \hfill
    \begin{subfigure}[t]{0.32\linewidth}
    		\centering
        \includegraphics[width=3.37cm]{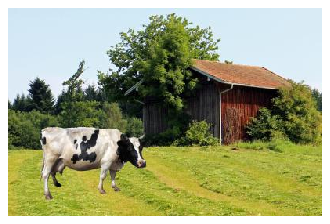}
        \caption{}
        \label{fig:cpgan:gen}
    \end{subfigure}
    \caption{{\bf Copy-pasting GAN.}
(a) The generator copies parts of
the `source' image and pastes them into the `destination' image,
such that the composite image looks real to the discriminator.
(b) If the generator copies a non-object (top),
over-segments (middle) or badly segments the object (bottom),
it is penalized by the discriminator.
(c) If the generator copies an object perfectly,
it is much harder for the discriminator to tell the image is fake.
}
    \label{fig:cpgan}
\end{figure}
}

\newcommand{\figAll}{
\begin{figure}[t]
    \centering
    \includegraphics[width=\textwidth]{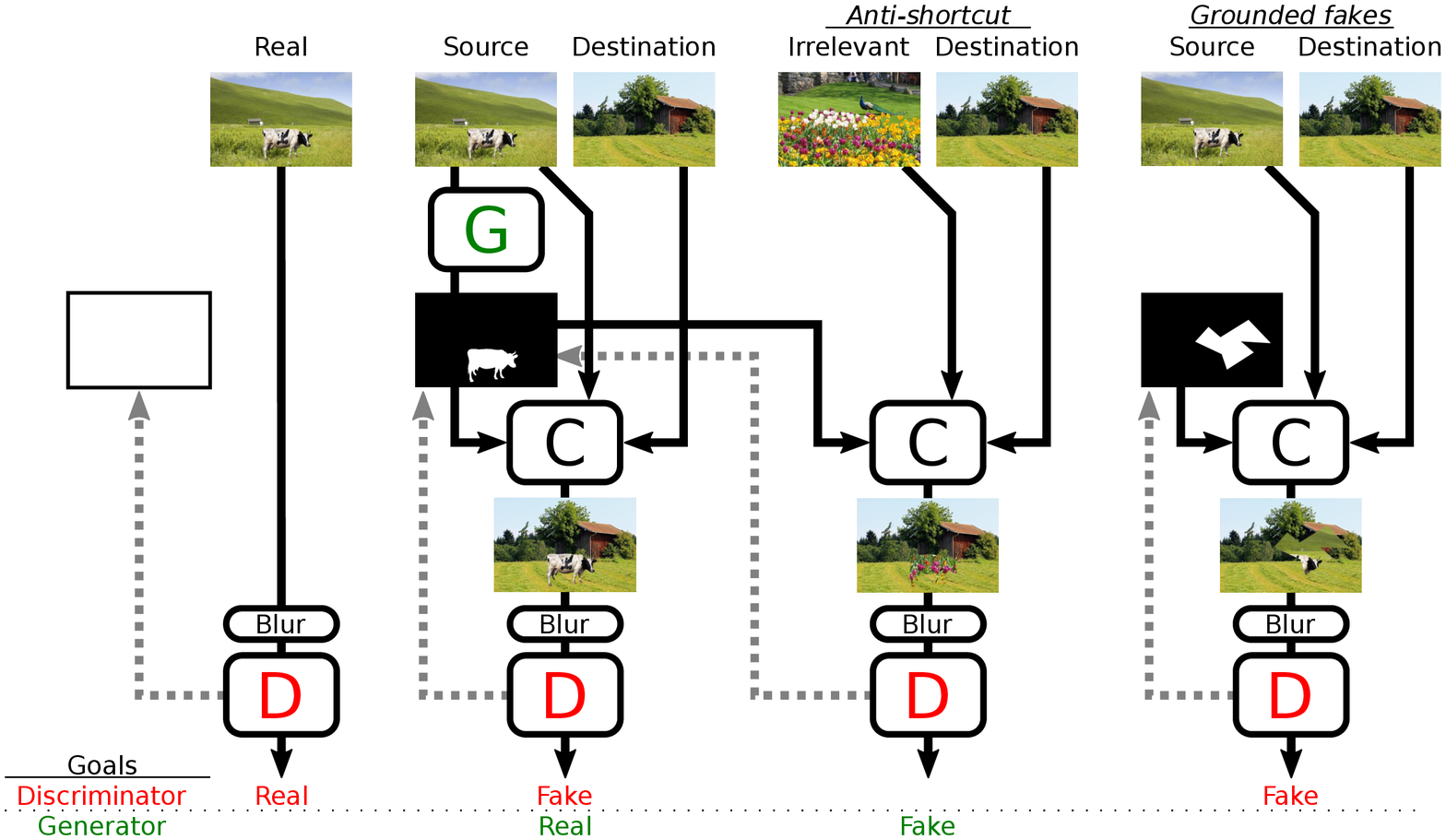}
    \caption{{\bf CP-GAN training.}
The {\bf \underline G}enerator creates a fake image by producing
a copy-mask from the source image and using it to blend the source
and the destination images together;
blending is performed using the {\bf \underline C}ompositing module
(\cf eq.\ \eqref{eq:c}).
Two leftmost columns: the standard GAN training where
the {\bf \underline D}iscriminator learns to distinguish between
real samples and generated ones,
while the {\bf \underline G}enerator tries to fool it.
The `Anti-shortcut' branch prevents the generator from
finding trivial solutions by enforcing that the generated mask
is not universally applicable, i.e.\ using it to blend  an irrelevant image
with the destination should produce a fake-looking image
(Section~\ref{sec:genshortcut}).
The `Grounded fakes' branch helps train the discriminator
by providing additional negatives (Section~\ref{sec:discguide}).
The grey dashed arrows illustrate the
`Auxiliary mask prediction' (Section~\ref{sec:discguide}).
`Border-zeroing' (Section~\ref{sec:genshortcut}) is not drawn for clarity.
All the {\bf \underline D}iscriminators are the same,
\ie they share weights.
}
    \label{fig:all}
\end{figure}
}

\newcommand{\figQual}{
\def\qualW{0.23\linewidth}
\begin{figure}[t]
    \centering
    \begin{subfigure}[t]{\qualW}
   		\centering
        \includegraphics[width=\textwidth]{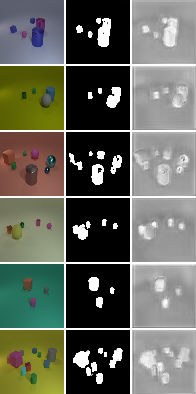}
        \caption{CLEVR+bg}
        \label{fig:qual:clevr}
    \end{subfigure}
    \hfill
    \begin{subfigure}[t]{\qualW}
  		\centering
        \includegraphics[width=\textwidth]{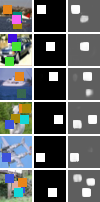}
        \caption{Squares}
        \label{fig:qual:squares}
    \end{subfigure}
    \hfill
    \begin{subfigure}[t]{\qualW}
  		\centering
        \includegraphics[width=\textwidth]{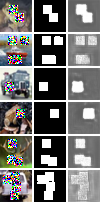}
        \caption{NoisySquares}
        \label{fig:qual:noisysquares}
    \end{subfigure}
    \hfill
    \begin{subfigure}[t]{\qualW}
  		\centering
        \includegraphics[width=\textwidth]{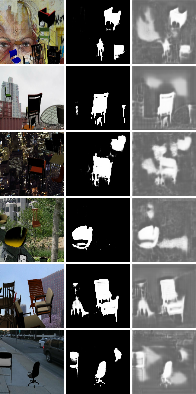}
        \caption{Flying Chairs}
        \label{fig:qual:flyingchairs}
    \end{subfigure}
    \caption{{\bf CP-GAN object discovery qualitative results.}
Columns show the
(left) input image,
(middle) discovered object (\ie copy-mask), and
(right) seediness.
}
    \label{fig:qual}
\vspace{-0.7cm}
\end{figure}
}

\newcommand{\tabRes}{
\begin{table}[t]
\centering
    \caption{{\bf Results and ablation studies.}
`Inst-C' stands for the `Instance Colouring' generator architecture.
The benefits of all training improvement techniques are investigated:
Anti-shortcut (Anti-sh.) and Border-zeroing (Border-z.)
from Section~\ref{sec:genshortcut},
Blur from Section~\ref{sec:discshortcut},
and
Grounded fakes (Grd-f.) and Mask prediction (M-p.)
from Section~\ref{sec:discguide}.
}
\begin{tabular}{@{~}ccccccrr@{~~~}r@{~~}c@{~~}rr@{~~}r@{}} \toprule
	\multicolumn{6}{c}{Variants} &
	\multicolumn{3}{c}{Squares} &
    ~ &
	\multicolumn{3}{c}{CLEVR+bg}
	\\
	\cmidrule{7-9} \cmidrule{11-13}
	Model &
	Anti-sh.\ & Border-z.\ & Blur & Grd-f.\ & M-p.\ &
	Mean & \multicolumn{1}{c}{Std} & Stab.\ &&
	Mean & \multicolumn{1}{c}{Std} & Stab.\
	\\ \midrule
    Inst-C &   &   &   &   &   &  1.6 & 1.7 &  20 &&  0.2 & 0.3 &  0 \\
    Inst-C & + &   &   &   &   &  1.6 & 1.1 &   0 && 13.9 & 30.1 &  0 \\
    Inst-C &   & + &   &   &   & 51.4 & 8.7 &   0 && 64.9 & 11.3 & 60 \\
    Inst-C & + & + &   &   &   & 95.6 & 1.0 & 100 && 97.6 & 0.9 & 30 \\
    Inst-C & + & + & + &   &   & 92.7 & 0.8 & 100 && 98.2 & 0.5 & 60 \\
    Inst-C & + & + &   & + &   & 98.0 & 0.6 & 100 && 96.8 & 2.1 & 20 \\
    Inst-C & + & + &   &   & + & 95.8 & 1.1 & 100 && {\bf 98.3} & 1.2 & 60 \\
    Inst-C & + & + &   & + & + & {\bf 98.3} & 0.3 & 100 && 96.9 & 0.9 & 50 \\
    Inst-C & + & + & + & + & + & 97.0 & 0.6 & 100 && 93.2 & 2.6 & {\bf 80} \\
    Direct & + & + &   & + & + & 95.1 & 0.4 & 100 && 60.7 & 3.7 & 40 \\
    Direct & + & + & + & + & + & 93.9 & 0.4 & 100 && 61.6 & 4.0 & 70 \\
    \bottomrule
\end{tabular}
\label{tab:res}
\end{table}
}

\newcommand{\tabComp}{
\centering
\begin{tabular}{@{}lr@{~~}r@{~~}r@{}} \toprule
	Methods & Sq. & NoisySq. & Fly.Ch.
	\\ \midrule
    CP-GAN &               98.3 & 100.0 & 40.3  \\
    MONet-top1 &            0.0 &  7.7 &   6.2 \\
    MONet-top2-oracle  &    1.1 &  7.7 &  34.4 \\
    MONet-top3-oracle  &   26.1 &  7.7 &  38.7 \\
    MONet-top4-oracle  &   73.0 &  7.7 &  39.7 \\
    MONet-all-oracle   &   99.6 & 52.3 &  40.8 \\
    \bottomrule
\end{tabular}
\label{tab:comp}
}

\newcommand{\figComp}{
\def\monetH{3.1cm}
    \centering
    \includegraphics[height=\monetH]{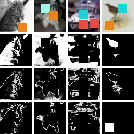}
    \hfill
    \includegraphics[height=\monetH]{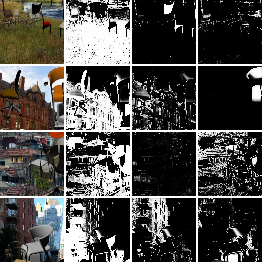}
}

\newcommand{\comp}{
\begin{figure}[t]
  \begin{minipage}[b]{0.45\linewidth}
    \centering
    \par\vspace{0pt}
    \tabComp
    \par\vspace{0pt}
  \end{minipage}%
  \hfill
  \begin{minipage}[b]{0.48\linewidth}
    \centering
    \par\vspace{0pt}
    \figComp
    \par\vspace{0pt}
\end{minipage}
\caption{{\bf Comparison to MONet~\cite{Burgess19}.}
ODP (left),
image and top 3 MONet slots for \emph{Squares} (middle) and
\emph{Flying Chairs}
(right).
Top slots focus on modelling the cluttered background,
while completely failing on
\emph{NoisySquares}.
CP-GAN handles these challenges well (Figure~\ref{fig:qual}).
}
\label{fig:comp}
\vspace{-0.5cm}
\end{figure}

}

\begin{abstract}
We tackle the problem of object discovery, where
objects are segmented for a given input image, and the system is trained
without using any direct supervision whatsoever.
A novel copy-pasting GAN framework is proposed, where the generator
learns to discover an object in one image by compositing it into another image  such that the discriminator cannot
tell that the resulting image is fake. After carefully addressing subtle
issues, such as preventing the generator from `cheating',
this game results in the generator learning to select
objects, as copy-pasting objects is most likely to fool the discriminator.
The system is shown to work well on four very different datasets,
including large object appearance variations in
challenging cluttered backgrounds.
\end{abstract}

\section{Introduction}
\label{sec:intro}

In this work, we consider the challenging task of unsupervised object discovery in 
images~\cite{Sivic05b,Russell06,Grauman06,Faktor14,Rubinstein13,Cho15,Vo19}.
Namely, the goal is to learn to detect and segment out prominent
objects in a scene, and train the system without any direct supervision on object detection or segmentation.
This model could then be used as pre-training for
supervised object detection with few annotations,
or as a vision module for an agent acting autonomously in an environment,
making it learn objects faster and more data-efficiently than if it
had to start from scratch with raw pixels as input.

We design a novel training procedure for object discovery,
taking inspiration from
generative adversarial networks (GANs) \cite{Goodfellow14a}.
The gist of the procedure is that the generator copies a region of one image
into another, and tries to fool the discriminator into thinking
the composite image is real. The intuition is that in order
to achieve this goal, the generator has to learn to discover and copy objects,
since the image will otherwise clearly look fake (Figure~\ref{fig:cpgan}).

However, there are multiple difficulties
in training this \emph{copy-pasting GAN (CP-GAN)}.
Namely, (i) the generator can take \emph{shortcuts} to `cheat'
and not perform the task as desired, (ii) 
the discriminator can use low-level cues instead of semantics,
and (iii)  GAN training is notoriously unstable.
We address all of these issues with novel solutions,
making the difference between the approach not working at all,
and a reasonable object discovery performance on synthetic
but challenging visual data.

An interesting distinguishing feature and a core novelty of the CP-GAN
compared to other object discovery approaches is that it is trained purely
\emph{discriminatively} --
the `generator' does not directly generate the object,
which can be hard due to having to model all aspects of appearance,
but rather detects and segments out an existing object.
This is unlike recent object discovery methods which do generate the object, 
rely on a pixel-based loss,
and are limited to simple scenes with uncluttered backgrounds~\cite{Eslami16,Greff17,Burgess19,Greff19}.
While generative methods are appealing due to the
\emph{``What I cannot create, I do not understand''} argument,
recent years have seen discriminative methods dominating
in image classification \cite{Krizhevsky12,He16},
object detection \cite{Girshick14,Redmon16} and
segmentation \cite{He17,Liu18}.
Furthermore, due to its discriminative nature,
it is in principle possible to use any object segmentation
network within the CP-GAN framework, potentially benefiting from
the latest state-of-the-art advances in the field.

\subsection{Related work}

\paragraph{Object discovery.}
Classical approaches to image-based unsupervised object discovery include
topic modelling \cite{Sivic05b,Russell06,Grauman06}
and clustering \cite{Faktor14},
often incorporating feature or part matching, geometric constraints and
saliency \cite{Rubinstein13,Cho15,Vo19}.
They all make use of hand-crafted features,
while some also incorporate saliency measures and (often supervised) object proposals.
Recent approaches based on deep learning take the inverse-graphics route
of creating a generative model of
the scene \cite{Eslami16,Greff17,Burgess19,Greff19}
and therefore tend to be limited to synthetic images with
uncluttered backgrounds.
In contrast, our approach, 
while still limited to synthetic imagery, can handle cluttered backgrounds coming from real images.
Other works have proposed to segment objects by interaction \cite{Pathak18}
or use motion as a strong cue for
object existence \cite{Papazoglou13,Zhang13,Yang19}.
Here we focus on image-based object discovery as our goal is to discover all
objects, regardless of whether they move or not.

\paragraph{Instance segmentation.}
Our generator network performs object instance segmentation,
and we draw inspiration from a large body of work on this topic,
which we very briefly summarize next.
Note however that all methods are fully supervised,
while our goal is unsupervised object discovery.
Most popular is the `propose and verify' approach, which perform
bottom-up object or mask proposals followed by a classification
and potentially mask prediction \cite{Pinheiro15,He17,Liu18}.
Another direction is `instance colouring',
where the network produces a dense feature map,
which are then either clustered to produce the instance segmentations \cite{Harley16,Kong18},
or seeds are chosen to grow the clusters
\cite{Fathi17,Brabandere17,Newell17,Novotny18}.

\paragraph{GANs for learning.}
Generative adversarial networks (GANs) \cite{Goodfellow14a}
are overwhelmingly being used for generating images,
such as full generation from scratch \cite{Radford16,Karras18,Brock19},
image translation \cite{Isola17,Zhu17,Bousmalis18},
superresolution \cite{Ledig17}, inpainting \cite{Demir18}
and denoising \cite{Zhong18b}.
Our approach is not concerned with the generating data aspect of GANs,
but uses them as a learnt loss function, a discriminative but
unsupervised training signal for object discovery.
Recently many works use adversarial loss functions
for tasks such as
learning domain-invariant representations \cite{Ganin15},
feature learning \cite{Radford16,Donahue17},
imitation learning \cite{Ho16},
human pose estimation \cite{Kanazawa18},
text recognition \cite{Gupta18a},
data augmentation \cite{Wang18,Dwibedi17,Tripathi19},
knowledge distillation \cite{Belagiannis18}, \etc.
GANs have also been used in translation \cite{Zhang17,Lample18,Conneau18},
and to learn to decipher coded texts \cite{Gomez18}.
No work has used GANs for object discovery.

\paragraph{GANs and copy-pasting.}
GANs have been used for copy-pasting \cite{Lee18,Lee19},
but the goal of these works was not to learn what to copy as we do,
but rather to create realistically looking visual data
for user-driven image \cite{Lin18,Wu17} and video \cite{Lee19} manipulation,
or data augmentation \cite{Dwibedi17}.
\cite{Lee18} learn where to paste an object, but require full supervision
and correct semantic segmentation as input to the network at test time.
The most related approach to ours is from Remez \etal \cite{Remez18},
who use a copy-pasting GAN for weakly supervised segmentation.
They use object category detectors, trained with full supervision,
and then employ a copy-pasting GAN within the predicted bounding boxes
to learn to segment the known objects.
Compared to \cite{Remez18},
we focus on completely unsupervised object discovery and segmentation,
propose a novel unsupervised method of avoiding trivial solutions
(Section~\ref{sec:genshortcut}),
two ways of increasing the training stability (Section~\ref{sec:discguide}),
and a segmentation method inspired by instance colouring
which has to be trained with reinforcement learning techniques
(Section~\ref{sec:genarch}).

\section{Copy-pasting GAN}

In this section, we introduce the copy-pasting GAN,
describe the motivation behind its design,
explain how to effectively train it,
and design the generator and discriminator network architectures
appropriately.

\subsection{Method}
\label{sec:method}

\figCpgan

The core idea behind the copy-pasting GAN (CP-GAN) is illustrated
in Figure~\ref{fig:cpgan:cpgan}.
As with a standard GAN, the generator and the discriminator play
a game where the discriminator's goal is to correctly classify
if a given image is real or fake,
while the generator's goal is to produce fake images which look
real to the discriminator.
The main difference from a standard GAN is in the way the generator
produces the fake image -- instead of directly outputting the pixels,
it combines two images, `source' and `destination',
by copying a region from the `source' into the `destination'.

The hypothesis motivating this design is that,
in order to fool the discriminator, the generator has to
copy objects, therefore achieving the unsupervised object discovery goal
of this work. If the generator copies a random non-object region,
or only a part of an object, or segments out the object badly,
the discriminator can easily deem the resulting image to be fake
(Figure~\ref{fig:cpgan:fake}).
However, if the generator selects the object perfectly,
the discriminator's task is much harder (Figure~\ref{fig:cpgan:gen}).

In this work, we restrict the generator --
the `source' image $I_s$ is passed to a ConvNet which produces
a pixel-wise segmentation mask, `copy-mask` $m_\theta(I_s)$;
all values of $m$ are in $[0, 1]$,  and $\theta$ are trainable parameters.
The composite image is produced by blending the `source' $I_s$
with the `destination' $I_d$ using the mask  $m_\theta(I_s)$:
\begin{equation}
	I_c = C\left(I_s, I_d, m_\theta(I_s)\right)
		= m_\theta(I_s) \odot I_s + \left( 1 - m_\theta(I_s)\right) \odot I_d
	\label{eq:c}
\end{equation}
\noindent  where
$\odot$ is the element-wise (Hadamard) product,
with a slight abuse of notation as the channel dimension is 
broadcast, \ie the same mask is used across the RGB channels.
This form of the generator limits the copy-paste operation such that
the region copied from source is pasted onto the same location
in the destination and no transformation is applied to it.
While somewhat restrictive, this is often a reasonable choice
if the data consists of images taken from similar camera poses
\eg self-driving car cameras, robotic environments, \etc.
In the future, it would be interesting to also learn where to paste,
and to apply geometric and photometric transformations to the pasted object.

\subsection{Crucial considerations and effective training}

While the core CP-GAN idea is simple and intuitive,
it is flawed due to having unwanted trivial solutions,
and training is quite unstable in practice;
in the following we address both of these issues.
The overview of the complete setup is shown in Figure~\ref{fig:all}.

\figAll

\subsubsection{Preventing shortcuts in the generator: Avoiding trivial solutions}
\label{sec:genshortcut}

While the hope behind the copy-pasting GAN is that the generator
has to learn to segment out objects in order to fool the discriminator,
the basic setup has a major flaw -- there are \emph{shortcuts} the
generator can take that do not require object recognition.
These trivial solutions have to be addressed in order to nudge the system
towards learning the intended behaviour.
In this section,
we describe two types of generator shortcuts and how to prevent them.

\paragraph{Copy one image in its entirety.}
Two simple strategies can make sure that the produced image looks
completely real -- the generator can completely copy the
source (`copy-all') or the destination (`copy-nothing')
image by producing constant copy masks
consisting of all-ones or all-zeros, respectively.
This is a major problem since it is very easy for a network to learn
to produce a constant output, and in practice
we find that the generator always degenerates to this behaviour when
there are no counter measures.
Furthermore, there are many variants of the above shortcuts, such as
copy-masks that are very sparse, or filled with non-zero but small values,
\etc.
Designing an additional loss which directly penalizes all such behaviours
is likely to be hard, tedious and fragile.
Instead, we opt for a more principled solution, described next.

The main unifying characteristic of these shortcuts is that the same
copy-mask can be applied blindly without taking the source image into
account at all.
One might then think that using a loss that maximizes the mutual information
between the source image and the copy-mask would solve the issue.
However, this can be `cheated' as well --
\eg setting copy-mask to $1/1000$ of the grayscale source image
would have a large mutual information while still effectively taking the
copy-nothing shortcut.
Instead, we devise a method that directly penalizes
\emph{all} `universally applicable' shortcut copy-masks,
without making any assumptions of whether the copy-mask depends on the
source image or not, whether the shortcut is a variant of copy-all,
copy-nothing or some other more sophisticated `cheat'.

This is accomplished by making use of the discriminator and another
\emph{irrelevant} image, $I_i$, randomly sampled from the dataset.
While the main objective of the CP-GAN is to
make the composition $C(I_s, I_d, m_\theta(I_s))$ look real to
the discriminator
(where $m_\theta(I_s)$ is generated from source $I_s$, \cf eq.\ \eqref{eq:c}),
we add an \emph{anti-shortcut} component to the generator loss which
penalizes real-looking $C(I_i, I_d, m_\theta(I_s))$,
as shown in Figure~\ref{fig:all}.
A `universally applicable' copy-mask used to composite an irrelevant image
(not used to produce the copy-mask)
and a destination image would by definition also look real,
which is directly discouraged by the loss.
Therefore, the generator is forced to learn to produce copy-masks
which are \emph{only} appropriate for the given source image,
and are inappropriate for other images.

Note that the proposed method is completely unsupervised,
as required for object discovery.
This is in contrast to Remez \etal \cite{Remez18} who prevent
these shortcuts by using supervision --
\eg to counter copy-nothing they make sure that the source object
is present in the composite image by running a check with
an object detector trained in a fully supervised manner.

\paragraph{Copy-all with a twist.}
There is one more shortcut, which bypasses the anti-shortcut branch
due to not being `universally applicable', and we describe it in detail in
\isArXiv{Appendix~\ref{supp:sec:genshortcut}.}{the supplementary material.}
To prevent it, we perform \emph{border-zeroing} --
the borders of the copy-mask are clamped to zero \cite{Remez18}.

\subsubsection{Preventing shortcuts in the discriminator: Avoiding easy real/fake discrimination}
\label{sec:discshortcut}

In order for the generator in the CP-GAN to learn about objects,
the discriminator has to provide it with a useful learning signal.
However, if the discriminator can latch on some very low-level cues
that the compositing image is not real, such as edge artefacts,
the GAN-game will overly focus on discovering and fixing small
(for this work) uninteresting tampering cues.
Therefore, we blur the input to the discriminator \cite{Sajjadi18}, to
hinder it from taking easy shortcuts.
In the future, it might be worth investigating better compositing
techniques such as Poisson blending \cite{Perez03} or \cite{Wu17},
as well as hiding other tampering cues \cite{Ye07,Lin09,Salloum18,Huh18,Zhou18}.

\subsubsection{Guiding the discriminator for increased stability}
\label{sec:discguide}

The previous section showed how to make sure the discriminator's job
is not too easy in order to give the generator a chance to learn.
This is not to say that the discriminator should be bad -- it also has to be
good enough in order to provide useful information for the generator.
In this section we consider two difficulties faced by the discriminator
in the copy-paste GAN setup and propose methods to alleviate them.

\paragraph{Jump-starting learning via grounded fakes.}
The anti-shortcut component (Section~\ref{sec:genshortcut})
makes an implicit assumption that the discriminator is good.
However, if the generator takes a copy-nothing shortcut
early in training, the discriminator never sees examples of truly fake images
as the generated images are actually real since they
are identical to the source.
This prevents the discriminator from learning anything meaningful,
and therefore training collapses.

To alleviate this issue, we make sure the discriminator does see fake images.
We artificially inject \emph{grounded fakes},
fake images created by randomly compositing two real images,
into the training loop and make sure the discriminator is able to tell
them apart from real images. In this way, there is one
branch that helps training the discriminator,
but which is completely independent of the generator and therefore
cannot be affected by it cheating, as shown in Figure~\ref{fig:all}.
Details of the generation of the `grounded fakes' are provided in
\isArXiv{Appendix~\ref{supp:sec:grounded}.}{the supplementary material.}

\paragraph{Auxiliary mask prediction.}
\label{sec:maskaux}
Training the discriminator effectively can be hard even when the generator
is not cheating because the real/fake labels contain too little information,
they do not tell the discriminator \emph{why} the generated image looks fake.
Some works \cite{Li16,Isola17} use fully-convolutional
discriminators which provide dense patch-level supervision.
However, this approach is not applicable for copy-pasting GANs because
most of the local patches will actually look real due to completely originating
from the source or the destination images.
Instead, we set up an auxiliary task where, apart from training the
discriminator on the standard image-level real/fake labels,
we also train it to predict the copy-mask used for compositing.
This task acts as additional dense supervision, and it points the
discriminator at the compositing evidence. The mask prediction
task is added to all branches: real (here the copy-mask is empty),
fake, anti-shortcut, and grounded fakes.
Details of the loss are in
\isArXiv{Appendix~\ref{supp:sec:losses}.}{the supplementary material.}

\subsection{Generator architectures}
\label{sec:genarch}

The generator network ingests the source image, $I_s$,
and produces the copy-mask
$m_\theta(I_s)$ which is then used to create the composite image
by blending the source and the destination (\cf eq.\ \eqref{eq:c}).
We explore two architectures based on the standard
U-Net \cite{Ronneberger15},
their core design is outlined next while full details
are given in
\isArXiv{Appendix~\ref{supp:sec:arch}.}{the supplementary material.}

\paragraph{Direct.}
The U-Net directly produces a 1-channel output representing
the copy-mask. The output of the final convolutional layer is
passed through a sigmoid in order to produce the mask in the
required $[0, 1]$ value range.

\paragraph{Instance colouring.}
Drawing inspiration from the instance segmentation method of \cite{Fathi17},
we design a network
to produce a dense pixel-wise $d$-dimensional feature map,
as well as pick a `seed' location.
These two induce a copy-mask by computing all similarities between
the seed-feature  and the pixel-wise feature map and passing these through a sigmoid.
Intuitively, the network could learn to assign a different `colour'
(\ie feature) to each object instance, and then pick a particular `colour'
to produce the copy-mask.
This can naturally extend to multi-object segmentation by iteratively
sampling new seeds, as opposed to the \emph{direct} architecture which produces
a single mask and would require a non-trivial extension in order to handle
multi-object segmentation.

The network is a U-Net which produces the feature map as well as
a seediness score (probability of picking a pixel as a seed).
Since at training time there is a discrete decision when picking the seed,
the network is trained using an actor-critic \cite{Sutton98} approach by making the analogy
with reinforcement learning -- the chosen seed is interpreted as the action taken,
and the seediness score is the policy,
and the negative generator loss is the reward.
Full details are available in
\isArXiv{Appendix~\ref{supp:sec:arch:gen}.}{the supplementary material.}

\paragraph{Related work.}
The main difference from \cite{Fathi17} is that they train
the network in a fully supervised manner, with straight-forward
supervised losses for both feature and the seed proposal heads.
In contrast, our instance colouring network is trained fully unsupervised
within the CP-GAN framework, requiring techniques from reinforcement learning (RL)
to train the seed proposal head.
Few other works train RL agents in an adversarial manner,
such as for imitation learning \cite{Ho16},
the painting agent of \cite{Ganin18},
and
the sequence generators of \cite{Yu17,Chen17a,Dai17}.

\subsection{Discriminator architecture}

The discriminator is also U-Net-based,
where the middle encoding is average-pooled and passed to
a fully connected layer to make the real/fake binary decision.
The decoder part of the U-Net is used to infer the underlying
copy-mask, as needed for the auxiliary mask prediction task
(Section~\ref{sec:maskaux}).
Full details are available in
\isArXiv{Appendix~\ref{supp:sec:arch:disc}.}{the supplementary material.}

\section{Experiments and discussions}

In this section we evaluate
CP-GAN's ability to discover objects, perform ablation studies, and
compare with a representative recent approach.
Full implementation details, including network architectures,
loss definitions, training schedule, \etc,
are provided in
\isArXiv{the appendices.}{the supplementary material.}

\paragraph{Datasets.}
Four datasets are used to evaluate object discovery (Figure~\ref{fig:qual}).
Ground truth segmentation masks for each object are available
but are only used for evaluation purposes, and are never seen
during training.

\emph{CLEVR+bg}: we use a modified CLEVR dataset \cite{Johnson17},
where we regenerate
all images by rendering the scenes with randomly coloured (one of 16 colours)
backgrounds, and resize them to $64 \times 64$. %
This is because we found the original CLEVR dataset to be
too simple as it contains only grey backgrounds, so object discovery can be performed trivially via a non-grey detector.

\emph{Squares}: we generate $32 \times 32$ images by
using CIFAR-10 \cite{Krizhevsky09} images as backgrounds and adding a random number
(between 1 and 5) of randomly coloured (one of 16 colours)
squares of length 9 pixels on top.
Although the `objects' (the squares) are simple, the backgrounds are
very cluttered.

\emph{NoisySquares}: as \emph{Squares}
but with large random salt and pepper noise added onto the squares.

\emph{Flying Chairs \cite{Dosovitskiy15}}: Publicly available dataset of
realistically rendered chairs overlaid on top of real cluttered backgrounds,
downscaled to $64 \times 64$.

\paragraph{Evaluation measures.}
As is standard in object segmentation, the quality of the segmentation mask
(\ie copy-mask)
is assessed using the intersection over union (IOU) score,
where a good match is deemed to be achieved if $IOU > 0.5$ \cite{Everingham10}.
However, evaluating in our fully unsupervised setting carries two difficulties:
(i) there is no control over which object the network picks, and
(ii) the network can select multiple objects at once.
Therefore, for a given test image,
we compute all valid ground-truth segmentation masks being
all possible combinations of masks induced by ground-truth objects,
\eg for objects masks $\{a, b, c\}$ all valid masks are 
$\{a, b, c, a \cup b, a \cup c, a \cup b \cup c\}$.
Object discovery for a test image is deemed successful if
the copy-mask achieves an $IOU > 0.5$ for any of the valid ground-truth masks
(\eg  the network discovering $a$, or discovering both  $a$ and $b$, are equally acceptable).
In practice, instead of explicitly computing all combinations,
greedy search is performed to find the best matching valid mask.
The final object discovery performance, ODP, is the percentage of successful
discoveries over the test set.

Each experiment is run 10 times, and we report the mean and standard deviation
of the early stopping performance.
We also measure the percentage of times that the training
does not eventually collapse,
\ie the ODP after the last training step
is at least $10\%$ of the maximal achieved performance.
This is a useful indication of the training stability, and reveals
the likelihood of achieving good performance when there is no way of
doing early stopping (\eg due to lack of held out labelled validation data).
However, note that clearly stability on its own is not a good measure,
as a method can be 100\% stable by being constantly bad,
so it should always be considered jointly with ODP scores.

\figQual

\paragraph{Absolute performance.}
Figure~\ref{fig:qual} shows representative examples of the CP-GAN
object discovery.
On most datasets, the method often selects and segments out objects
correctly, achieving ODP of 98.3\%, 100\% and 98.3\% for
\emph{Squares}, \emph{NoisySquares} and \emph{CLEVR+bg}, respectively.
This confirms the main hypothesis behind CP-GAN -- in order to fool
the discriminator, the generator naturally learns to discover objects.
On the most realistic images with a relatively large variation in object
appearance and cluttered backgrounds, \emph{Flying Chairs},
CP-GAN achieves promising results (40.3\%).

\paragraph{Number of discovered objects.}
CP-GAN tends to select all objects at once, apart from on \emph{Squares}
where it picks single objects.
Note however that for the latter, the generator does often
find multiple objects, as evidenced by the seediness score, but it only selects
one -- this could be used to segment out multiple objects by iteratively sampling
seeds.
We observe generators select fewer objects as training progresses,
presumably because this reduces the amount of possible edge artefacts
the discriminator can find. It is also possible that the discriminator could
make an informed guess on whether an image is real or generated based on the number
of objects present in it, which could nudge the generator towards
selecting single objects.

\paragraph{Generator architectures.}
While both the \emph{Direct} and the \emph{Instance Colouring}
generator architectures (Table~\ref{tab:res}) achieve a reasonable performance,
Inst-C wins, especially on CLEVR+bg (98.3\% vs 61.6\%).

\tabRes

\paragraph{Ablation studies.}
Table~\ref{tab:res} shows various combinations of the methods proposed
to prevent the generator and discriminator from cheating, and to stabilizing
the training (Sections~\ref{sec:genshortcut}, \ref{sec:discshortcut} and
\ref{sec:discguide}).
As expected,
it is critical to always use both \emph{anti-shortcut} and \emph{border-zeroing}
-- disabling one or both makes the generator learn to take unwanted shortcuts,
while enabling them achieves
a reasonable object discovery performance.
Guiding the discriminator through the \emph{auxiliary mask prediction} task
consistently improves performance and
stability.
Providing an additional source of
\emph{grounded fakes} significantly improves results on \emph{Squares},
but is slightly detrimental on \emph{CLEVR+bg}.
\emph{Blurring} the input to the discriminator helps preventing it from taking
shortcuts and therefore stabilizes training, while slightly decreasing
the quality of segmentation masks as the generator is not punished enough for being
too sloppy.
This is not evident on \emph{Squares} since training on this dataset is
always stable when \emph{anti-shortcut} and \emph{border-zeroing} are included.
However, it makes a large difference on \emph{CLEVR+bg} where
\emph{blurring} improves the stability by 30\% percentage points
in all three configurations.

\comp

\paragraph{Generative vs discriminative methods.}
We compare CP-GAN to the recent generative object discovery approach
MONet~\cite{Burgess19}
which iteratively decomposes the scene into multiple `slots', the top slots
being most important.
The ODP scores are computed for the top slot (`top1'),
and cases where an `oracle' picks per-example best slot out of top K (`topK-oracle`)
and all slots (`all-oracle').
Figure~\ref{fig:comp} shows MONet's performance and illustrates
the problems affecting all generative models trained with
pixel-level losses:
(i) strong focus
on explaining the cluttered background, \eg the oracle for top 3 slots
achieves only 26.1\% ODP on \emph{Squares}, and
(ii) complete failure in the presence of very varied objects
(\emph{NoisySquares}).
Good discriminative methods such as CP-GAN are able to ignore the background
and handle large appearance variations as detection is easier than
reconstruction.

\section{Conclusions}

We have shown that unsupervised object discovery can be achieved
by training the copy-pasting GAN if special care is taken
to prevent the generator from learning to take shortcuts.
The approach shows promising results on four varied datasets with significant
background clutter, which has plagued recent methods.
This is a first step towards real-world object discovery, for which one
might need to scale up, use deeper networks, train with a curriculum,
learn where to paste, or apply geometric and photometric transformations.
Furthermore, the CP-GAN framework has potential beyond object discovery,
as to compile a completely realistic image,
the generator should learn where to place objects such that
they are compatible with
scene geometry, physical properties, object relations, \etc.

\clearpage
{\small
\bibliographystyle{ieee}
\bibliography{bib/shortstrings,bib/vgg_local,bib/vgg_other,bib/more}
}

\appendix
\newcommand{\figCheats}{
\def\cheatsW{0.24\linewidth}
\def\cheatsWI{0.83\textwidth}
\def\raiseH{1.2cm}
\def\spaceW{\hspace*{0.007\linewidth}}
\def\corrspace{\phantom{$I_\text{g}$}}
\begin{figure}[t]
    \centering
    \begin{subfigure}[t]{\cheatsW}
    		\centering
        \mbox{
        \raisebox{\raiseH}{\rotatebox[origin=t]{90}{\phantom{$I_d$}}}
        \includegraphics[width=\cheatsWI]{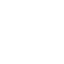}  %
        }
        \mbox{
        \raisebox{\raiseH}{\rotatebox[origin=t]{90}{$I_s$}}
        \includegraphics[width=\cheatsWI]{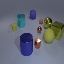}
        }
        \mbox{
        \raisebox{\raiseH}{\rotatebox[origin=t]{90}{$I_d$}}
        \includegraphics[width=\cheatsWI]{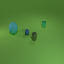}
        }
        \mbox{
        \raisebox{\raiseH}{\rotatebox[origin=t]{90}{$I_i$}}
        \includegraphics[width=\cheatsWI]{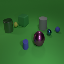}
        }
        \caption{Input images}
    \end{subfigure}
    \spaceW
    \begin{subfigure}[t]{\cheatsW}
    		\centering
        \mbox{
        \raisebox{\raiseH}{\rotatebox[origin=t]{90}{copy-mask\corrspace}}
        \includegraphics[width=\cheatsWI]{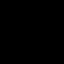}
        }
        \mbox{
        \raisebox{\raiseH}{\rotatebox[origin=t]{90}{$I_c$\corrspace}}
        \includegraphics[width=\cheatsWI]{figures/clevr_cheat_dest.png}
        }
        \mbox{
        \raisebox{\raiseH}{\rotatebox[origin=t]{90}{$I_\text{anti\_shortcut}$\corrspace}}
        \includegraphics[width=\cheatsWI]{figures/clevr_cheat_dest.png}
        }
        \mbox{
        \raisebox{\raiseH}{\rotatebox[origin=t]{90}{$I_\text{border\_zeroing}$\corrspace}}
        \includegraphics[width=\cheatsWI]{figures/clevr_cheat_dest.png}
        }
        \caption{Copy-nothing}
    \end{subfigure}
    \spaceW
    \begin{subfigure}[t]{\cheatsW}
    		\centering
        \mbox{
        \raisebox{\raiseH}{\rotatebox[origin=t]{90}{copy-mask\corrspace}}
        \includegraphics[width=\cheatsWI]{figures/clevr_cheat_copyall.png}
        }
        \mbox{
        \raisebox{\raiseH}{\rotatebox[origin=t]{90}{$I_c$\corrspace}}
        \includegraphics[width=\cheatsWI]{figures/clevr_cheat_source.png}
        }
        \mbox{
        \raisebox{\raiseH}{\rotatebox[origin=t]{90}{$I_\text{anti\_shortcut}$\corrspace}}
        \includegraphics[width=\cheatsWI]{figures/clevr_cheat_irr.png}
        }
        \mbox{
        \raisebox{\raiseH}{\rotatebox[origin=t]{90}{$I_\text{border\_zeroing}$\corrspace}}
        \includegraphics[width=\cheatsWI]{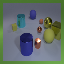}
        }
        \caption{Copy-all}
    \end{subfigure}
    \spaceW
    \begin{subfigure}[t]{\cheatsW}
    		\centering
        \mbox{
        \raisebox{\raiseH}{\rotatebox[origin=t]{90}{copy-mask\corrspace}}
        \includegraphics[width=\cheatsWI]{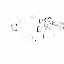}
        }
        \mbox{
        \raisebox{\raiseH}{\rotatebox[origin=t]{90}{$I_c$\corrspace}}
        \includegraphics[width=\cheatsWI]{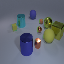}
        }
        \mbox{
        \raisebox{\raiseH}{\rotatebox[origin=t]{90}{$I_\text{anti\_shortcut}$\corrspace}}
        \includegraphics[width=\cheatsWI]{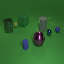}
        }
        \mbox{
        \raisebox{\raiseH}{\rotatebox[origin=t]{90}{$I_\text{border\_zeroing}$\corrspace}}
        \includegraphics[width=\cheatsWI]{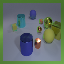}
        }
        \caption{Smart copy-all}
        \label{supp:fig:cheats:copyalltwist}
    \end{subfigure}
    \caption{{\bf Generator shortcuts.}
(a) `Source', `destination' and `irrelevant' input images.
(b) Copy-nothing shortcut where the copy-mask is made of all 0s,
the composite image equals the destination.
(c) Copy-all shortcut where the copy-mask is made of all 1s,
the composite image equals the source.
(d) Copy-all with a twist, where the copy-mask is made of mostly 1s
but with some 0s aligned with source edges, making the composite
image visually similar to the source.
Anti-shortcut loss prevents (b)-(c)
as using the same copy-mask on an irrelevant image also yields a real-looking
image, which is penalized.
However, it does not help with (d) as $I_\text{anti\_shortcut}$ also looks
fake since the copy-mask has 0s aligned with $I_s$ edges -- these
artefacts are hard to see in $I_c$ but stand out in $I_\text{anti\_shortcut}$
(\eg see artefacts to the right of the grey cylinder;
best viewed in digital form).
Border-zeroing addresses (c)-(d) as it creates a large border artefact
for copy-all like shortcuts, but does not address (b) as it has no effect
on copy-nothing.
Therefore, both anti-shortcut and border-zeroing are needed to counter
all generator shortcuts, as verified by
\isArXiv{Table~\ref{tab:res}.}{Table~1 of the main paper.}
}
    \label{supp:fig:cheats}
\end{figure}
}

\newcommand{\figUNet}{
\begin{figure}[t]
    \centering
    \includegraphics[width=\textwidth]{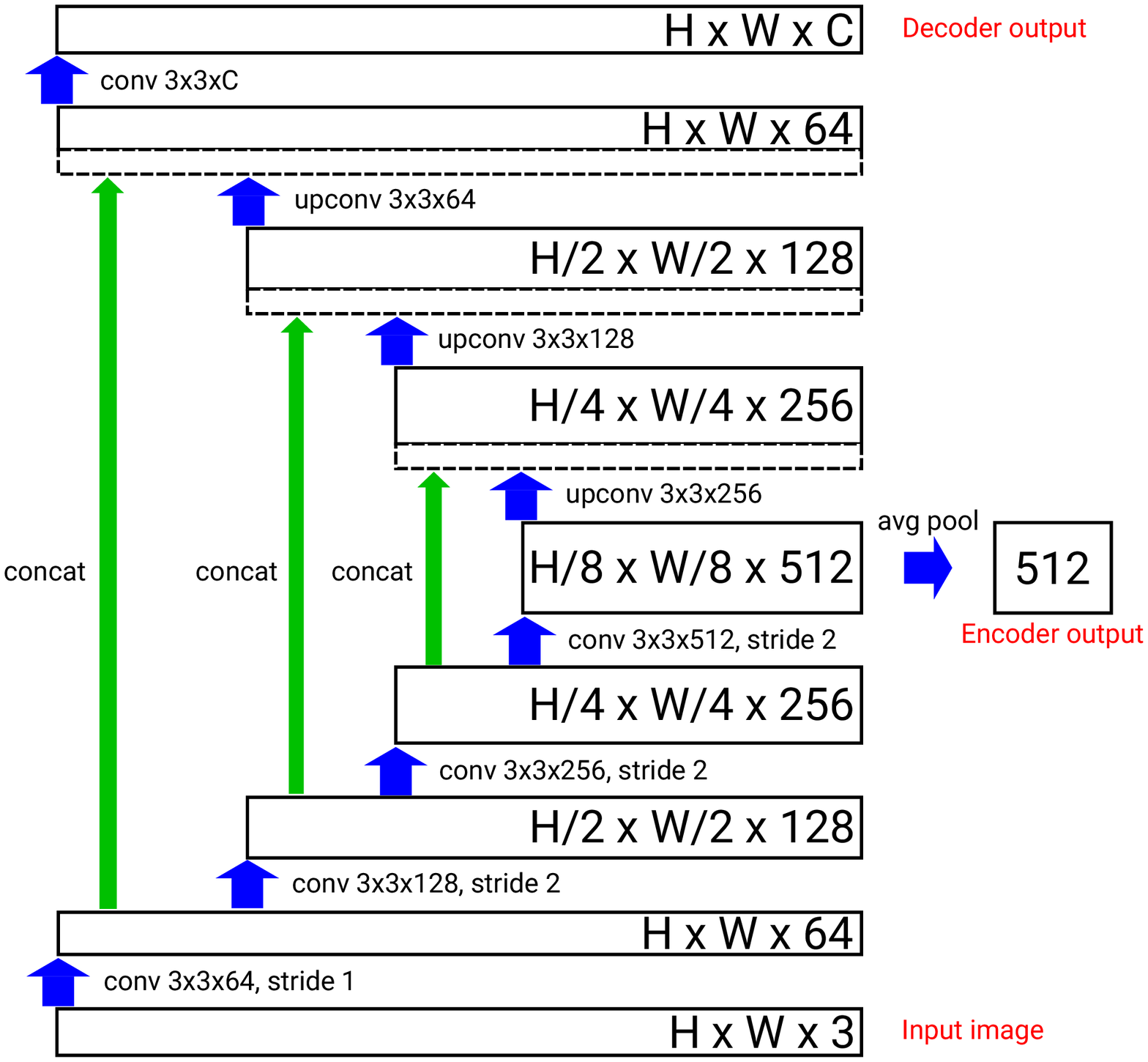}
    \caption{{\bf U-Net building block.}
Black rectangles represent tensors whose dimensions are specified
by the inner text, arrows are operations in the neural network.
Given an $H \times W$ input image, the network produces two outputs
-- a pixel-wise $C$-dimensional output (`decoder output'),
and a 512-D vector (`encoder output').
All convolutions and upconvolutions apart from the top one are
followed by instance normalization \cite{Ulyanov17} and LeakyReLU nonlinearity.
Following \cite{Odena16},
upconvolutions are implemented as upsampling by a factor of 2
with nearest neighbour interpolation,
followed by a standard $3 \times 3$ convolution with stride 1.
}
    \label{supp:fig:unet}
\end{figure}
}

\isArXiv{
\section*{Overview of the appendices}
}{
\section*{Overview of the supplementary material}
}

We provide all details which were excluded from the main paper
\isArXiv{for clarity,}{due to length constraints,}
but are not essential for understanding the core
ideas and contributions of the work.
This includes
the exact definition of the losses used to train CP-GAN,
the training procedure for the Instance Colouring network,
additional examples of the generator shortcuts,
details of the generation of `grounded fakes',
full specification of the network architectures,
as well as miscellaneous implementation details such as
the learning rate schedule.

\isArXiv{
}{
\tableofcontents
\newpage

}

\section{Losses}
\label{supp:sec:losses}

In this section provide the details of all the losses used to
train the CP-GAN. Their visual overview is available in
\isArXiv{Figure~\ref{fig:all}.}
{Figure~2 of the main paper.}

\paragraph{Setup and notation.}
Let the source and destination images be $I_s$ and $I_d$, respectively.
The discriminator $D$ (with parameters $\omega$)
judges whether an image $I$ looks real or fake by
producing a scalar $D(I)$ in $[0, 1]$, with larger values corresponding to
more realistic images.
The generator $G$ (with parameters $\theta$)
produces a composite image from two input images
$I_c = G(I_s, I_d)$. As explained in
\isArXiv{Section~\ref{sec:method},}{the main paper,}
here we consider
a specific form of the generator where, given the source image $I_s$,
it produces a copy-mask $m_\theta(I_s)$
which is then used to composite the two images as follows:

\begin{equation}
	I_c = G(I_s, I_d) = C\left(I_s, I_d, m_\theta(I_s)\right)
		= m_\theta(I_s) \odot I_s + \left( 1 - m_\theta(I_s)\right) \odot I_d
    \label{supp:eq:c}
\end{equation}

\noindent  where
$\odot$ is the element-wise (Hadamard) product,
with a slight abuse of notation as the channel dimension is 
broadcast, \ie the same mask is used across the RGB channels.

Let us denote the cross entropy loss for
prediction $p$ and label $l$, both in $[0, 1]$,
as:

\begin{equation}
\loss(x, l) = -l \log p - (1-l) \log(1 - p)
\end{equation}

\paragraph{Standard GAN losses.}
As per standard,
the discriminator's goal is to classify a real image $I_r$ as being real
and generator produced image $I_c$ as fake,
while the generator is trying to fool the discriminator into
classifying $I_c$ as real;
the following cross entropy losses reflect this GAN game.

The discriminator loss for a real image, $I_r$, is:

\begin{equation}
\loss_{D_\text{real}} = \loss\left(D(I_r), 0.75\right)
\end{equation}
\noindent where,
following best practice \cite{Goodfellow16} to decrease the discriminator overfitting,
label smoothing \cite{Szegedy16} is performed only for this loss
(\ie instead of the desired label being $1$ it is $0.75$).

The discriminator loss for a generator-produced fake image, $I_c$, is:

\begin{equation}
\loss_{D_\text{fake}} = \loss\left(D(I_c), 0\right)
\end{equation}

The generator loss is the negative of the discriminator loss for the
fake image:

\begin{equation}
\loss_{G_\text{fake}} = -\loss\left(D(I_c), 0\right)
\end{equation}

\noindent where $I_c$ is produced as in eq.~\eqref{supp:eq:c} and therefore
depends on the generator parameters $\theta$.

\paragraph{Anti-shortcut.}
As described in
\isArXiv{Section~\ref{sec:genshortcut},}{the main paper,}
the generator should not
produce an `universally applicable' copy-mask $m_\theta(I_s)$.
The following loss penalizes the generator if using $m_\theta(I_s)$ to composite
an irrelevant image, $I_i$, with $I_d$ yields a real-looking image
$I_\text{anti\_shortcut}$:

\begin{eqnarray}
	I_\text{anti\_shortcut} = C\left(I_i, I_d, m_\theta(I_s)\right) \\
\loss_{G_\text{anti\_shortcut}} = \loss\left(D(I_\text{anti\_shortcut}), 0\right)
\end{eqnarray}

\paragraph{Grounded fakes.}
Given a `grounded fake' image $I_\text{grounded\_fake}$,
the discriminator's goal is to
classify it as being fake:

\begin{equation}
\loss_{D_\text{grounded\_fake}} = \loss\left(D(I_\text{grounded\_fake}), 0\right)
\end{equation}

\paragraph{Auxiliary mask prediction.}
As an auxiliary task, the discriminator is also stimulated to predict
the copy-mask used for creating the composite image (input to the discriminator).
The discriminator therefore, apart from producing the score $D(I)$,
also produces a copy-mask estimate $m_\omega(I)$.
Note that there is an inherent ambiguity in the copy-mask
prediction, since $C(I_1, I_2, m)$ produces
the same result as $C(I_2, I_1, 1-m)$;
since the discriminator does not have access to either of the original images,
it cannot know if it should output $m$ or $1-m$.
Therefore, we use a permutation invariant loss \cite{Yu17a} for this auxiliary task:

\begin{equation}
\loss_\text{mask}(m_\omega(I), m) = \min\left(
    \loss\left(m_\omega(I), m\right),
    \loss\left(m_\omega(I), 1-m\right)
    \right)
\end{equation}

\noindent with the slight abuse of notation as here the cross entropy loss 
$\loss$ is applied between two images (instead of scalars),
and is equal to the average pixel-wise cross-entropy loss.

The auxiliary losses are added to all the branches:
real (here the copy-mask is empty), fake, anti-shortcut, and grounded fakes;
these are equal to:

\begin{eqnarray}
\loss_{\text{mask}_\text{real}} & = & \loss_\text{mask}(m_\omega(I_r), 0) \\
\loss_{\text{mask}_\text{fake}} & = & \loss_\text{mask}(m_\omega(I_c), m_\theta(I_s)) \\
\loss_{\text{mask}_\text{anti\_shortcut}} & = & \loss_\text{mask}(m_\omega(I_\text{anti\_shortcut}), m_\theta(I_s)) \\
\loss_{\text{mask}_\text{grounded\_fakes}} & = & \loss_\text{mask}(m_\omega(I_\text{grounded\_fakes}), m_\text{grounded\_fakes})
\end{eqnarray}

\noindent where $m_\text{grounded\_fakes}$ is the random mask used to
composite two images to create the `grounded fakes' image
(see Section~\ref{supp:sec:grounded} for details).

The total auxiliary loss is the sum of the above:

\begin{equation}
\loss_\text{aux} =
\loss_{\text{mask}_\text{real}} +
\loss_{\text{mask}_\text{fake}} +
\loss_{\text{mask}_\text{anti\_shortcut}} +
\loss_{\text{mask}_\text{grounded\_fakes}} 
\end{equation}

\paragraph{Final losses.}
The above losses are added up to form the generator and discriminator
losses used to train the CP-GAN.
We have not experimented with tuning the relative loss weights.

The total generator loss is:

\begin{equation}
\loss_G = \loss_{G_\text{fake}} + \loss_{G_\text{anti\_shortcut}}
\end{equation}

The total discriminator loss is:

\begin{equation}
\loss_D = \loss_{D_\text{real}} + \loss_{D_\text{fake}} +
\loss_{D_\text{grounded\_fakes}} + 0.1 \loss_\text{aux}
\end{equation}

\section{Instance Colouring: Further details and training}
\label{supp:sec:ic}

Recall from
\isArXiv{Section~\ref{sec:genarch}}{Section 2.3 of the main paper}
that the Instance Colouring
generator functions as follows:
given an input image $I_s$, it produces dense pixel-wise features $f_\theta(I_s)$
as well as pixel-wise seediness score $s_\theta(I_s)$.
Dropping $I_s$ for clarity,
let the feature and seediness at pixel $(i, j)$ be $f_\theta^{ij}$ and $s_\theta^{ij}$,
respectively.
For a given seed $(y,x)$, the value of the induced copy-mask at pixel $(i,j)$,
$m_\theta^{ij}$ is:

\begin{equation}
m_\theta^{ij} = sigmoid({f_\theta^{yx}}^T f_\theta^{ij})
\end{equation}

At test time, $(y,x)$ are picked as $\argmax_{i,j} s_\theta^{ij}$.

To train the network, we make an analogy with reinforcement learning --
picking the seed location is an action ($a=(i,j)$)
and the corresponding reward ($r_\theta^a$)
is the negative generator loss resulting from using the induced mask $m_\theta^a$
to make the composite image.
The seediness score is interpreted as the policy,
so the entire system can be trained with a policy gradient method
such as REINFORCE \cite{Williams92}.

In more detail, we wish to maximise the expected reward under the policy:

\begin{equation}
E_{a \sim s_\theta} [r_\theta^a], ~~ \text{where } r_\theta^a \equiv -\loss_G(m_\theta^a)
\end{equation}

Note that unlike the standard reinforcement learning setting,
there are two interacting paths towards increasing the expected reward --
by improving the policy (\ie the seed choice via the seediness score)
as well as by increasing the reward itself (\ie by learning better
features which together with the seed induce the copy-mask
that fools the discriminator more). To maximise with gradient descent,
we compute the derivatives with respect to the generator parameters $\theta$:

\begin{eqnarray}
\nabla_\theta E_{a \sim s_\theta} [r_\theta^a] =
\nabla_\theta \sum_{i,j} s_\theta^{ij} r_\theta^{ij} =
\sum_{i,j} s_\theta^{ij} (\nabla_\theta r_\theta^{ij}) +
\sum_{i,j} (\nabla_\theta s_\theta^{ij}) r_\theta^{ij}
\\
=
E_{a \sim s_\theta}[\nabla_\theta r_\theta^a] +
(\nabla_\theta E_{a \sim s_\theta}) [r_\theta^a]
\end{eqnarray}

The REINFORCE trick \cite{Williams92} can be applied to
the second term to obtain the gradient
of the expectation with respect to the policy, yielding:

\begin{equation}
\nabla_\theta E_{a \sim s_\theta} [r_\theta^a] =
E_{a \sim s_\theta}[
\nabla_\theta r_\theta^a +
r_\theta^a \nabla_\theta \log s_\theta(a)
]
\end{equation}

\noindent which, as with a standard policy gradient method,
is estimated using Monte Carlo sampling,
\ie by sampling an action (seed location) according to the policy (seediness score).
Furthermore, we use two standard improvements over REINFORCE:
(i) an additional entropy regularization term with 0.01 weight, and
(ii) variance reduction by using an actor-critic method \cite{Sutton98}
(\ie  as per standard --
a separate network head predicts the value of the action,
which is subtracted from the reward and not backpropagated through,
and is trained with square loss to approximate the true reward).

\section{Generator shortcuts}
\label{supp:sec:genshortcut}

Here we give more details on generator shortcuts, complementing
\isArXiv{Section~\ref{sec:genshortcut}.}{Section 2.2.1 of the main paper.}
Examples of shortcuts and counter measures
are shown in Figure~\ref{supp:fig:cheats}.

\figCheats

\paragraph{Copy-all with a twist.}
While the anti-shortcut method is very effective
at suppressing generator shortcuts, we found one additional type of
cheat that the generator was still able to find, and address it next.
Namely, the generator can perform an imperfect copy-all strategy,
where the copy-mask consists of all ones apart from some zeros
placed on prominent edges in the source image.
This has the effect of looking fairly real as the result is
the source image with some edges being slightly imperfect, which
is not easy to detect (Figure~\ref{supp:fig:cheats:copyalltwist})
and can therefore fool the discriminator.
However, this cunningly bypasses the anti-shortcut branch since
the resulting copy-mask is not `universally applicable' as the
imperfections are aligned with source edges.
Used when compositing with an irrelevant image, the edges stand out
and the outcome looks fake, satisfying the anti-shortcut loss.

To prevent this behaviour, we adapt the copy-all shortcut
prevention mechanism of \cite{Remez18}. Namely, the borders of the copy-mask
are clamped to zero, so that any version of copy-all produces images
with strange looking borders (Figure~\ref{supp:fig:cheats}),
and are therefore penalized by the discriminator;
we name this method \emph{border-zeroing}.

\section{Generation of grounded fakes}
\label{supp:sec:grounded}

The `grounded fakes'
\isArXiv{Section~\ref{sec:discguide}}{(Section 2.2.3 of the main paper)}
are created by randomly compositing two images.
We use the same source and destination images as the main branch,
but instead of the copy-mask
produced by the generator, we use a randomly generated copy-mask.
The random mask is produced by sampling a random polygon and
filling the inside with 1s and outside with 0s.
The random polygon sampling is conducted as follows
(all numbers are expressed in normalized image coordinates).
First, a central point is sampled inside the
rectangle with corners (0.1, 0.1) and (0.9, 0.9).
Then, the number of vertices $v$ is sampled uniformly from $[4, 6]$.
The polar coordinates, with respect to the polygon centre,
of each of the $v$ vertices is obtained
by sampling the radius in $[0.1, 0.5]$ and angle in $[0, 2\pi]$.

All of the parameters have been chosen arbitrarily to produce reasonable
looking polygons.
They have not been optimized as we don't believe their exact values have
a large impact on CP-GAN training.

\section{Architectures}
\label{supp:sec:arch}

All network architectures are built on top of a
U-Net \cite{Ronneberger15} style network.
The basic building block shown in Figure~\ref{supp:fig:unet}
is reused in all network architectures
-- the two generator types and the discriminator --
as described next.

\figUNet

\subsection{Generator}
\label{supp:sec:arch:gen}

\paragraph{Direct.}
The direct architecture produces a single copy-mask for a given image.
This is done by using the U-Net block (Figure~\ref{supp:fig:unet})
with $C=1$, where the decoder output is followed by a sigmoid
to produce the copy-mask.

\paragraph{Instance colouring.}
As described in
\isArXiv{Section~\ref{sec:genarch}}{the main paper,}
the instance colouring architecture
produces a pixel-wise feature map as well as a pixel-wise seediness score.
It also produces pixel-wise value estimates (Section~\ref{supp:sec:ic})
used to train the network.
This is accomplished by using the U-Net block (Figure~\ref{supp:fig:unet})
with $C=66$ -- the pixel-wise 66-dimensional decoder output
is interpreted as a concatenation of
64-D features, 1-D seediness logits and 1-D value estimates.
The seediness logits are passed through a softmax as required
to produce the seediness score, being the probability of picking a location
as the seed.

\subsection{Discriminator}
\label{supp:sec:arch:disc}

The discriminator network produces a single scalar predicting whether
the input image is real or not. Additionally, for auxiliary mask prediction,
it also outputs pixel-wise copy-mask prediction.
This is done by using the U-Net block (Figure~\ref{supp:fig:unet})
with $C=1$
-- the decoder output is passed through a sigmoid to produce
the copy-mask estimate,
while the encoder output is passed through a single fully connected layer
followed by a sigmoid to yield the realness score.

\section{Miscellaneous implementation details}
\label{supp:sec:misc}

\paragraph{Training details.}
CP-GAN has been trained using the Adam optimizer \cite{Kingma15}
with default parameters and no weight decay.
The first 1K training steps (batches) we only train the discriminator,
following which we alternate
1 step of generator with 1 step of discriminator training.
For \emph{Squares} and \emph{NoisySquares} we use one GPU
and batch size 256, with an initial learning rate of $3 \cdot 10^{-4}$.
For \emph{CLEVR+bg} and \emph{Flying Chairs} we use 4 GPUs
with a per-GPU batch size of 80 and synchronous training, with
an initial learning rate of $10^{-4}$.
The learning rate is divided by $3$ after 30K training steps.
We terminate the training after 300K steps, although typically good object
discovery performance is achieved much earlier
(\eg 90\%+ for \emph{Squares} after 20K steps, or peak performance for
\emph{Flying Chairs} at 150K steps),
but we keep training in order to test the stability.
\isArXiv{Method was implemented with
TensorFlow \cite{tensorflow},
Sonnet \cite{sonnet} and 
TF-Replicator \cite{Buchlovsky19} libraries.}{}

Recall, the seediness score produced by the Instance Colouring network
is the probability for each position in the image to be picked
as the seed.
A potential outcome of training the network
is that the seediness score is high only for a single object.
To discourage this, structured dropout is performed on the seediness scores
during training -- we randomly sample a square of size $1/3$ of
the minimal image dimension, and set the seediness scores inside the square
to zero. This effectively means that the dropout can randomly enforce
the generator to not pick an object it intended to, so it has to pick an
alternative by sampling a seed from outside the square region. If the seediness
focuses on a single object, it will incur a loss when this object is vetoed by
the dropout. Therefore, the optimal strategy is for the seediness to contain all
objects.
The dropout is not essential as CP-GAN works well without it too, but it
encourages full object coverage.

For blurring we use a Gaussian kernel with $\sigma=1$ and kernel size $3$.

\paragraph{MONet~\cite{Burgess19} comparisons.}
For comparisons with MONet~\cite{Burgess19} we perform a sweep
over the various hyperparameters --
learning rate, number of slots, $\sigma_{bg}$, $\sigma_{fg}$ and $\gamma$.

\paragraph{Flying Chairs \cite{Dosovitskiy15} ground truth.}
The dataset does not actually
contain ground truth segmentations. However, it does provide ground truth
optical flow between frames, which we use to infer the object segmentations.
We exploit the fact that all the movement of all objects and the background
are governed by the underlying affine transformations used to generate
the two frames. For each $3 \times 3$ patch we fit an affine transformation
consistent with the ground truth flow, keep only the consistent ones,
and cluster the affine transformations to obtain the objects and their
segmentations. After some postprocessing such as removal of
tiny objects (noise) and the largest one (background),
we obtain fairly good quality ground truth segmentations.

\end{document}